# EHSAN: Leveraging ChatGPT in a Hybrid Framework for Arabic Aspect-Based Sentiment Analysis in Healthcare

EMAN ALAMOUDI*

School of Computing, Newcastle University, Newcastle, UK, E.S.O.Alamoudi2@newcastle.ac.uk

ELLIS SOLAIMAN

School of Computing, Newcastle University, Newcastle, UK, Ellis.Solaiman@newcastle.ac.uk

**Abstract**. Arabic-language patient feedback remains under-analysed because dialect diversity and scarce aspect-level sentiment labels hinder automated assessment. To address this gap, we introduce EHSAN, a data-centric hybrid pipeline that merges ChatGPT pseudo-labelling with targeted human review to build the first explainable Arabic aspect-based sentiment dataset for healthcare. Each sentence is annotated with an aspect and sentiment label (positive, negative, or neutral), forming a pioneering Arabic dataset aligned with healthcare themes, with ChatGPT-generated rationales provided for each label to enhance transparency. To evaluate the impact of annotation quality on model performance, we created three versions of the training data: a fully supervised set with all labels reviewed by humans, a semi-supervised set with 50% human review, and an unsupervised set with only machine-generated labels. We fine-tuned two transformer models on these datasets for both aspect and sentiment classification. Experimental results show that our Arabic-specific model achieved high accuracy even with minimal human supervision, reflecting only a minor performance drop when using ChatGPT-only labels. Reducing the number of aspect classes notably improved classification metrics across the board. These findings demonstrate an effective, scalable approach to Arabic aspect-based sentiment analysis (SA) in healthcare, combining large language model annotation with human expertise to produce a robust and explainable dataset. Future directions include generalisation across hospitals, prompt refinement, and interpretable data-driven modelling.

## 1 INTRODUCTION

Patient feedback is increasingly recognised as a vital metric for evaluating and improving healthcare quality. Unlike standardised surveys, which often limit expression, free-form reviews, such as online comments and social media posts, allow patients to share more detailed and emotionally nuanced accounts of their experiences [1]. These narratives have proven valuable for identifying service strengths and weaknesses, complementing structured satisfaction tools, and informing healthcare [2–4]. The healthcare industry itself acknowledges this shift. According to a recent estimate, 90% of healthcare leaders view patient experience as a top strategic priority, while 45% identify improving satisfaction scores as a key goal – often through digital technologies that support more responsive, patient-centred care [5]. This growing interest has accelerated the adoption of natural language processing (NLP) tools, enabling large-scale analysis of narrative feedback and helping extract actionable insights for quality improvement. Among these tools, sentiment analysis (SA) has emerged as a central technique, categorising text based on emotional polarity – typically positive, negative, or neutral. In healthcare, SA supports the automatic classification of patient opinions, helping to detect patterns of satisfaction and dissatisfaction [2]. However, the same study found that SA in healthcare is less developed than in domains such as retail, primarily due to the complexity of medical narratives and the scarcity of annotated datasets. To address these limitations, many health systems rely on structured instruments such as HCAHPS in the US [6], and the Saudi Patient Experience Measurement Program in Saudi Arabia [7]. These tools, while useful, often constrain patient responses. For instance, patients may over-report their satisfaction due to gratitude bias, which can result in an overestimation of the actual quality of care [8]. In contrast, unsolicited narrative feedback can offer richer insights into care experiences, driving researchers toward more open-ended, patient-driven sources. In Arabic-language contexts, however, progress remains limited due to dialectal diversity, morphological complexity, and a scarcity of annotated data. Overcoming these challenges is crucial for developing inclusive NLP tools that can support quality healthcare analysis across diverse linguistic settings.

Our study addresses these gaps by introducing a fine-grained, sentence-level Arabic healthcare review analysis. In summary, the key contributions of this study are as follows:

- **Fine-Grained Sentence-Level Aspect-Based Sentiment Analysis (ABSA)**. Arabic hospital reviews are segmented into individual sentences, each annotated with a specific service aspect and its corresponding sentiment. This enables SA at multiple levels, including both the document level and the more detailed aspect level within the text.
- **Hybrid Annotation Framework**. We introduce a three-tier annotation strategy combining ChatGPT pseudo-labelling with human review, producing fully supervised, semi-supervised, and unsupervised training sets. This design enables us to evaluate the impact of human oversight on model performance.
- **Explainable Annotations**. Each automated annotation is accompanied by a rationale generated by ChatGPT, improving transparency and allowing verifiability of the model's decisions.
- **Dual Taxonomy Levels**. We define two aspect category schemes – an initial 17-category taxonomy and a consolidated 6-category schema – to examine the effect of granularity on performance and to provide both detailed and high-level analysis options.
- **Empirical Evaluation with Transformer Models**. We implemented our explainable healthcare sentiment annotation (EHSAN) dataset and report comprehensive experiments fine-tuning an Arabic-specific BERT model (AraBERT) and a lightweight multilingual model (DistilBERT) on our data under different supervision levels and classification granularities. This provides insights into how a domain-specific model compares to a general-purpose model on Arabic ABSA, as well as how much manual annotation is needed for high accuracy.
- **Validation of Pseudo-Labelling in Low-Resource NLP**. Our results empirically demonstrate that high-quality pseudo-labels from a Large Language Model (LLM) can serve as a reliable alternative to manual labels in a low-resource setting, with minimal loss of accuracy. This highlights a cost-effective path for building language-specific resources when expert annotation is scarce.

- **Data-Management Perspective**. We articulate how schema evolution, provenance capture and privacy-aware indexing turn ABSA into a data-engineering problem of direct interest to the IDEAS (International Conference on Intelligent Data Engineering and Automated Systems) community.

The remainder of this paper is organised as follows. The literature review examines existing research on SA in the healthcare sector. The methodology section describes the processes used for data collection, preprocessing, and topic modelling of Arabic healthcare reviews, as well as the details of the annotation workflow using ChatGPT, including prompt design and tiered annotation, and the details of the training models used, along with the evaluation metrics applied to assess their performance. The results section presents the main findings, which are further interpreted in the discussion section. Finally, the conclusion and future work section summarises key insights and proposes directions for future research.

## 2  LITERATURE REVIEW

**SA in Healthcare**. Sentiment analysis (SA) has been applied in healthcare to automatically gauge patient satisfaction from text. However, as [2] found, healthcare SA lags behind other domains due to limited training data and the complexity of medical narratives. Traditional patient experience measures, including the HCAHPS survey in the United States [6], and the Saudi Patient Experience Measurement Program [7], typically quantify satisfaction but provide limited qualitative insight. In contrast, unsolicited feedback (e.g., online reviews, social media) can capture rich, emotionally charged experiences. For example, [9] analysed thousands of Hungarian health forum posts to identify common complaints (waiting times, communication issues, etc.). Such studies underscore the value of narrative data for uncovering issues not evident in surveys. However, challenges in processing this data include casual language, misspellings, and mixed sentiments.

**Aspect-Based Sentiment Analysis (ABSA)**. ABSA extends SA by linking sentiments to specific aspects or topics. In healthcare, this means identifying which part of the service a sentiment refers to (e.g., nursing staff, cleanliness, billing, etc.). [10] applied ABSA to 30,000 Indian hospital reviews, extracting aspect-specific sentiments (doctors, staff, facilities, cost) and showed that such granularity provides actionable insights for hospital management. In terms of the Arab world, research is nascent. [11] built a Saudi patient comments classifier (PX_BERT) that is able to categorise feedback into predefined aspect categories, but they did not perform sentiment labelling. [4] created datasets from Saudi Twitter and Google reviews using ChatGPT with human verification to annotate aspects such as medical staff, appointments, and the like, along with sentiment. Their HEAR dataset segmented feedback by aspect, but still operated on whole reviews (multi-label per review) rather than true sentence-level segmentation. Table 1 presents a summary of the Arabic datasets used in previous healthcare SA studies, along with the dataset used in the present study.

Table 1: Overview of Arabic datasets used in healthcare sentiment analysis studies

| Study | Dataset Name and Size | Granularity | Annotation Details | Annotation Rationale | Public Availability |
|---|---|---|---|---|---|
| [11] | PX dataset: labeled comments: 19,000, negative-only subset: 13,000 | Comment-level (multi-label classification over 25 SHCT categories; no sentiment) | Manually labelled by Saudi Ministry of Health (MOH) | No | No – Internal MOH dataset |
| [4] | HoPE-SA: 12,400 tweets and HEAR: 25,156 reviews | Aspect-level (up to 5 predefined aspects per review; each assigned an individual sentiment) | ChatGPT + human verification | No | Yes |
| Present study | EHSAN, 6000 reviews | Sentence-level: Reviews segmented into sentences; each sentence labelled with one aspect (from 17 fine-grained or 6 coarse-grained categories) and sentiment. Aspect-level information is derived through aggregation across sentences. | ChatGPT + human verification | Yes | Yes |

**LLMs for Annotation**. The advent of powerful LLMs, such as ChatGPT, has opened up new avenues for dataset creation in low-resource languages. Consequently, researchers have begun using LLMs to generate labels or augment data. For example, [12] applied few-shot prompting with ChatGPT to label emotion in French texts, achieving an F1 of 0.66 and outperforming classical classifiers. [13] used ChatGPT-4o to analyse Chinese patient reviews with high accuracy (F1 = 0.912), highlighting its ability to understand specialised and non-English texts. In Arabic healthcare NLP, [14] applied few-shot prompting with ChatGPT on a small Arabic SA corpus, reporting promising results and calling for a more comprehensive evaluation across tasks and datasets.

Despite these advances, challenges remain. While previous studies have introduced Arabic healthcare review datasets, none have provided the level of detail found in this work, particularly in terms of the granularity of aspect categories and the inclusion of explanations generated by large language models (LLMs). Furthermore, concerns about the content generated by LLMs, such as potential bias or inaccuracies, necessitate careful validation. This study addresses these gaps by developing the EHSAN dataset, which incorporates a rigorous annotation scheme and evaluates the extent to which manual correction was needed to train effective models.

## 3 METHODOLOGY

A multi-stage pipeline was developed using ChatGPT to pseudo-label sentences in Arabic hospital reviews with aspect categories and sentiment, supported by varying levels of human review. This approach enables the capture of issue-specific sentiments within individual reviews and ensures alignment with Saudi Arabia's healthcare complaint taxonomy. The outcome was the explainable healthcare sentiment annotation (EHSAN), a dataset that provides explainable, multi-level annotations for Arabic patient feedback.

### 3.1 Data Collection and Description

A broad dataset of patient reviews was collected from the Google Maps listings of hospitals in Saudi Arabia. The dataset comprised reviews of hospitals across various regions of the country, reflecting diverse regional dialects and vocabularies. Each hospital entry included a list of reviews containing text, star ratings, and metadata. Specifically, reviews were chosen from three representative hospitals – one each from the central region (the city of Riyadh), the western region (the city of Jeddah), and the eastern region (the city of Dammam) regions – to ensure geographic and dialectal diversity. The three hospitals also differed in their average patient ratings (ranging from approximately 2.4 to 4.0 out of 5), thereby capturing a range of both positive and negative patient experiences. All three hospitals selected were officially accredited and listed in the national registry valid through 2025 [15].

A total of 5,000 reviews were included at this stage, selected from the three target hospitals. Following sentence segmentation (a process described in a later section), the reviews yielded 9,337 individual sentences. From this pool, 2,000 sentences were randomly selected from each hospital, resulting in a total of 6,000 sentences used for detailed analysis.

All data were collected in adherence to ethical standards. Only publicly available review text was used, and any personal identifiers were either absent or removed to protect reviewer privacy. The final dataset (the explainable healthcare sentiment annotation, or EHSAN) was constructed to support aspect-based SA at the sentence level, with each sentence annotated with a specific service aspect and its corresponding sentiment. Additionally, the dataset supports document-level analysis by including an overall sentiment label for each complete review, allowing for flexible use across various analytical scopes.

The study protocol was reviewed and exempted by Newcastle University Ethics Committee (Ref: 59967/2023). It also fully complies with Google Maps Terms of Service.

### 3.2 Data Pre-processing

The raw review texts, as collected, were found to contain various inconsistencies and noise typical of user-generated content, particularly in Arabic. To address these issues, a two-phase pre-processing pipeline was implemented to clean and standardise the text prior to analysis, as explained below.

**Structural Clean-Up.** In the first phase, attention was given to text segmentation and formatting. Regular expressions were applied to insert explicit sentence breaks at punctuation marks (e.g., periods, question marks, and exclamation points), ensuring that each sentence appeared on a separate line. This step was essential due to the informal nature of many Arabic reviews, which often lack clear sentence boundaries. Emojis were removed and replaced with either placeholders or newlines to prevent interference with text processing. Excess whitespace was normalised by collapsing multiple newlines and spaces, thereby promoting formatting consistency. Any empty or null entries were discarded following this cleaning step.

**Linguistic Normalisation.** The second phase focused on addressing Arabic-specific textual variations. The AraBERT Pre-processor toolkit from AUBMindLab was employed to standardise the Arabic text. Key transformations included the removal of diacritics (Tashkeel), normalisation of variant ya letters (e.g., converting forms like "ي" to the standard "ى"), and unification of other character variants to ensure consistent spelling. Punctuation that did not serve a syntactic purpose was eliminated, and sequences of repeated letters were reduced to manageable lengths (e.g., "مررررررحبا" was shortened to "مرحبا") to address exaggeration patterns typical of informal writing. Similarly, the decorative elongation of letters was removed. These steps, informed by best practices in Arabic NLP, were applied to reduce noise and enable the model to focus on the core content.

Following pre-processing, the text was rendered significantly cleaner and more standardised while maintaining the original meaning of the reviews. This thorough process ensured that the subsequent segmentation and labelling stages could be applied consistently, despite variability in writing styles across the dataset.

### 3.3 Topic Modelling and Sentence Segmentation

One novel aspect of the methodology used in the present study is employing topic-driven sentence segmentation performed by ChatGPT (specifically, the GPT-4o-mini model). Instead of relying solely on punctuation-based splitting – which can be unreliable for informal text – ChatGPT was guided to segment each review such that each resulting sentence corresponded to a single topical idea or aspect. This was achieved by prompting ChatGPT with tailored instructions to divide the review into meaningful sentences, allowing for the merging or splitting of phrases as needed to isolate distinct topics. When a sentence in the original text addressed multiple topics, it was broken down into separate sentences, each focusing on a single topic. Conversely, when multiple short sentences pertained to the same topic, they were merged to enhance coherence. This approach ensured that each segmented sentence could later be assigned a single relevant aspect category with accuracy.

To inform and support this process, topic modelling was conducted on the corpus as an analytical tool – used not directly in the labelling pipeline but to validate topic coverage. BERTopic, an advanced topic modelling technique, was employed, which is capable of clustering semantically similar texts and extracting representative keywords for each cluster. When applied to the pre-processed reviews (prior to segmentation), BERTopic identified approximately 25 distinct topics covering themes such as staff attitude, waiting times, and facility cleanliness. These unsupervised insights were strongly aligned with the predefined taxonomy of 17 aspect categories, thereby providing validation for the comprehensiveness of the chosen taxonomy.

For comparative purposes, a rule-based Arabic NLP tool, Stanza [16], was used for sentence splitting. It was observed that ChatGPT's segmentation yielded higher topical coherence, while Stanza occasionally retained compound sentences or split text at every punctuation mark, regardless of semantic context. Based on these findings, the segmentation produced by ChatGPT was adopted for the final dataset. Each segmented sentence was then passed on to the annotation stage.

### 3.4 Annotation Strategy with ChatGPT (Pseudo-Labelling)

Following segmentation, each sentence – now treated as an individual review – was annotated by ChatGPT with an aspect category and sentiment label using a few-shot learning approach. The prompt provided to ChatGPT included detailed instructions along with definitions of the aspect categories and sentiment labels. For each review, two labels were requested: (1) the main topic or aspect of the review, selected from a predefined list of 17 categories (e.g., "Medical_and_Nursing_Staff," "Billing_and_Finance"), and (2) the sentiment expressed – positive, negative, or neutral.

Additionally, ChatGPT was instructed to provide a brief justification for each classification decision, explaining the rationale behind its choice of topic and sentiment based on the content of the sentence. This justification was initially generated in English because the model demonstrated higher reliability in producing accurate reasoning in English. Subsequently, a translation of the justification into Arabic was requested.

To enhance clarity and analytical precision, each sentence generated by ChatGPT was treated as an independent review. Nevertheless, the original review IDs were preserved and linked to all extracted sentences to ensure traceability and facilitate flexible future use. This structured design allows for analysis at multiple levels – sentence, aspect, or document – and ensures the adaptability of the dataset.

Human validation was applied across all dataset splits—namely, the training, validation, and test sets. The training data were fully reviewed and categorized into three supervision levels: fully verified, partially reviewed, and model-labelled without correction. Human annotators were native Arabic speakers, all with higher education degrees, understanding of the healthcare context, and familiarity with digital review environments. All annotators participated in a structured training programme prior to the annotation task, which included a workshop on the 17-aspect taxonomy and hands-on exercises using a pre-labelled subset of 50 sentences. To ensure annotation quality, they were required to pass a qualification task before contributing to the dataset. The level of inter-annotator agreement was high, with disagreements occurring in only three instances, all of which involved borderline or ambiguous cases. Annotation was performed by two primary annotators working collaboratively, with data entered into a structured Excel sheet designed to review and evaluate ChatGPT's auto-generated labels. In cases of disagreement, a third reviewer adjudicated the final label using a majority voting protocol. In addition, a subset of model justifications was reviewed and found clear and consistent with final labels, reflecting strong alignment with the model's rationale. To evaluate the reliability of this manual annotation protocol, a stratified random sample of 600 sentences, constituting 5% of the 12,000 total annotations (6,000 for topics and 6,000 for sentiments), was selected for inter-annotator agreement analysis. Prior to adjudication, each sentence was independently annotated by the two primary reviewers. Interrater agreement was quantified using Cohen's Kappa (κ) [17], which yielded a score of 0.873, thus reflecting almost perfect agreement. Nevertheless, human reviews were incorporated to varying degrees, as described in subsequent sections. Table 2 presents the accuracy of ChatGPT in classifying topics and sentiments across the training, validation, and test sets, with all results verified through human annotation.

Table 2: ChatGPT accuracy in topic and sentiment classification

| Dataset | Topic | Sentiment | Total Cases |
|---|---|---|---|
| Training | 0.90 | 0.96 | 3600 |
| Validation | 0.93 | 0.96 | 1200 |
| Testing | 0.90 | 0.94 | 1200 |

### 3.5 Dataset Construction: Supervision Levels

A core component of the methodology involved evaluating varying levels of human involvement in the labelling process. To facilitate this, three versions of the training dataset were constructed:

**Fully Supervised Dataset (FSD).** All labels generated by ChatGPT (both aspect and sentiment) were fully reviewed and corrected by human annotators. This version represents a gold standard dataset in which every label was verified and considered accurate.

**Semi-Supervised Dataset (SSD).** In this version, half of the reviews in the training data were reviewed and corrected by human annotators, while the remaining half retained the original labels assigned by ChatGPT without modification. The human-reviewed portion was selected randomly to ensure that all aspect categories were represented and that no class bias was introduced.

**Unsupervised Dataset (USD).** No human corrections were applied in this version; the dataset relied entirely on ChatGPT's pseudo-labels. This variant was used to assess the model's performance when trained solely on machine-generated annotations.

The dataset was partitioned into three subsets: 3,600 reviews were allocated for training, 1,200 reviews were reserved for validation, and the remaining 1,200 were allocated for testing. The training subset was used as the foundation for generating the three versions described above, and each was subjected to a different level of human supervision. In contrast, all reviews within the validation and test subsets were fully annotated by human reviewers to ensure the presence of reliable ground truths for model evaluation. Due to the presence of English text in some reviews, any review in which English characters exceeded 25% of the total content was excluded. As a result, the final number of reviews in each subset became 3,583 for training, 1,189 for validation, and 1,190 for testing.

Additionally, two classification schemes were implemented for aspect categories within each dataset – one using the original 17 fine-grained classes, and another using six broader, consolidated classes. The 6-class scheme was developed by grouping semantically related fine-grained aspects, particularly where overlaps were observed. For instance, individual departments, such as radiology and surgery, were grouped under a unified "medical services" category. This grouping was introduced after it was observed that certain categories within the 17-class taxonomy were underrepresented (e.g., only four reviews pertained to privacy), and that even human annotators occasionally confused similar categories.

The six consolidated aspect categories were defined as medical services, administrative services, appointment and waiting, environment and facilities, billing and finance, and miscellaneous. This dual dataset structure – incorporating both 17-class and 6-class schemes – was designed to examine how classification granularity impacts model performance while also offering a more practical schema for applications in which high-level categorisation is sufficient.

### 3.6 Model Training and Evaluation

The annotated datasets were evaluated by fine-tuning two transformer-based models to perform multi-class topic classification and SA:

**AraBERT.** A BERT-language model pre-trained specifically on large Arabic corpora [18]. We used the AraBERT v0.2 (large) model, which has demonstrated strong performance on various Arabic NLP tasks and is well suited to handling both Modern Standard Arabic and dialectal Arabic text. We expected AraBERT to have an advantage in understanding the nuances of Arabic patient reviews.

**DistilBERT (multilingual).** A distilled (lightweight) version of BERT that supports multiple languages [19]. We chose the base multilingual DistilBERT to represent a more computationally efficient model. While it is not specialised in Arabic, it covers Arabic in its training and is much smaller than AraBERT. The comparison between AraBERT and DistilBERT allowed us to discern the trade-off between a large Arabic-specific model and a smaller general-purpose model for this task.

The classification models were fine-tuned using open-source frameworks, primarily Hugging Face and PyTorch, and all experiments were conducted on Paperspace using high-performance GPUs to accelerate training. We treated **aspect category classification** (17-class or 6-class) and **sentiment classification** (3-class) as two separate tasks. In practice, we trained each model for aspect classification and for sentiment classification, rather than a single multi-task model. This was to simplify training and because the aspect taxonomy changed (17 to 6) whereas sentiment remained the same. Each model was trained on the training set (FSD, SSD, or USD) accordingly, and evaluated on the common fully supervised test data. We used the same data splits for both models to ensure comparability.

**Training Details.** Standard fine-tuning procedures were applied. A batch size of 4 and a learning rate of 1e-5 were used, and fine-tuning was conducted for 5 epochs in each run; these hyperparameters were tuned on the validation set. Each model's own tokenizer was ensured: AraBERT's tokenizer was used for Arabic text, and DistilBERT's multilingual tokenizer was used for that model. Appropriate padding and truncation were applied to ensure that the input sequences conformed to the models' length requirements. Early stopping on the validation set was employed to prevent overfitting.

For evaluation, the macro-averaged F1 score was primarily reported, along with accuracy, macro-precision, and macro-recall for completeness. Macro-averaging (i.e., averaging metrics equally across all classes) was deemed appropriate due to class imbalance, ensuring that performance on rare classes was accounted for. Emphasis was placed on the F1 score, as it balances precision and recall – particularly important in a multi-class imbalanced scenario. Accuracy alone could be misleading if the model disproportionately favours majority classes. Training time (in minutes) was also measured for each model in each dataset to assess computational efficiency.

## 4 RESULTS

After the models were trained, their performance was analysed on the held-out test set. The results for the aspect (topic) classification task under both the 17-class and the 6-class schemes are presented below, followed by the sentiment classification results. All results were reported based on the fully human-annotated test set to ensure fairness in the evaluation.

### 4.1 Aspect Classification Performance (17 Classes)

Table 3 summarises the performance of AraBERT and DistilBERT on the 17-category aspect classification task for each training dataset variant (FSD, SSD, and USD). Both models achieved reasonably good accuracy and F1, given the difficulty of the task (17 imbalanced classes). AraBERT outperformed DistilBERT in all scenarios. For instance, with fully supervised training data, AraBERT reached 79% accuracy and an F1 score of 0.66, compared to DistilBERT's 72% accuracy and 0.61 F1. The performance gap persisted in semi-supervised and unsupervised settings, indicating the benefit of a language-specific model for this task. Importantly, the difference between using fully human-reviewed data and relying solely on ChatGPT labels was insignificant, with AraBERT's F1 dropping from 0.66 (FSD) to 0.64 (USD), and DistilBERT's falling from 0.61 to 0.59. The semi-supervised case was intermediate. Minor improvements were seen with human corrections, especially for DistilBERT (which improved by about 0.02 in F1 with full supervision). Overall, the FSD yielded the best performance for both models, followed by the SSD, and the USD (machine-only) was only slightly behind. Notably, the AraBERT model trained on the USD (pure ChatGPT labels) even outperformed the DistilBERT model trained on the FSD (fully human labels).

To assess the reliability of the model results and mitigate the effect of random variation, a two-step statistical analysis was conducted. First, 95% confidence intervals were estimated via bootstrapping to evaluate the stability of F1 scores. Second, an Approximate Randomization Test was done to assess the statistical significance of performance differences. Based on five independent runs with varied random seeds, the results showed consistent confidence intervals in most cases, except for AraBERT on the USD dataset, which exhibited greater variability ([0.46, 0.65]). In contrast, its confidence interval on the FSD dataset was narrow ([0.66, 0.68]), indicating stable and reliable performance. Statistically significant differences were observed only between the FSD and USD settings in AraBERT ($p = 0.021$), while other comparisons showed no significant differences. Statistical testing was applied to the 17-aspect setup due to observable variability, unlike the 6-aspect and sentiment results, which showed near-identical scores.

### 4.2 Aspect Classification Performance (Six Classes)

After merging the aspect labels into six broader categories, the model performance improved markedly. Table 4 shows the results of the 6-class classification. With fewer classes and more training examples per class (since we combined several related classes), the models achieved higher scores across all metrics. AraBERT, in particular, reached 81% accuracy and 0.78 F1 on the FSD, a substantial increase over the 17-class scenario. DistilBERT also improved to approximately 75% accuracy, to 0.69 F1 in the FSD case. Strikingly, AraBERT's performance was identical for the FSD and the SSD in the 6-class setup – 0.78 F1 in both cases. Even in the USD scenario (no human labels), AraBERT maintained an F1 of around 0.77. DistilBERT showed a slight drop from the FSD (0.69 F1) to the SSD (0.67) to the USD (0.67). These results affirm that reducing the label noise (via human review) has diminishing returns once the classes are made easier (6 instead of 17) – the pseudo-labels were good enough that cleaning half or all of them gave minimal additional benefit, at least for the stronger model.

Table 3: Aspect classification results (17 classes)

| Dataset | Model | Accuracy | Macro Precision | Macro Recall | Macro F1 Score | Confidence Interval 95% |
|---|---|---|---|---|---|---|
| FSD | AraBERT | 0.79 | 0.65 | 0.69 | 0.66 | [0.66, 0.68] |
| | DistilBERT | 0.72 | 0.60 | 0.62 | 0.61 | [0.61, 0.61] |
| SSD | AraBERT | 0.80 | 0.65 | 0.69 | 0.65 | [0.64, 0.67] |
| | DistilBERT | 0.71 | 0.60 | 0.61 | 0.60 | [0.60, 0.61] |
| USD | AraBERT | 0.78 | 0.64 | 0.68 | 0.64 | [0.46, 0.65] |
| | DistilBERT | 0.71 | 0.59 | 0.61 | 0.59 | [0.59, 0.61] |

Table 4: Aspect classification results (6 classes)

| Dataset | Model | Accuracy | Macro Precision | Macro Recall | Macro F1 Score |
|---|---|---|---|---|---|
| FSD | AraBERT | 0.81 | 0.78 | 0.78 | 0.78 |
| FSD | DistilBERT | 0.75 | 0.69 | 0.68 | 0.69 |
| SSD | AraBERT | 0.81 | 0.78 | 0.78 | 0.78 |
| SSD | DistilBERT | 0.73 | 0.68 | 0.67 | 0.67 |
| USD | AraBERT | 0.81 | 0.79 | 0.77 | 0.77 |
| USD | DistilBERT | 0.73 | 0.69 | 0.67 | 0.67 |

### 4.3 Sentiment Classification Performance

In addition to aspect prediction, sentiment classification (positive, negative, or neutral) was also performed for each review within the framework. Separate sentiment classifiers were trained on the same training splits. Strong performance was demonstrated by both AraBERT and DistilBERT, with AraBERT consistently achieving superior results across all datasets. On the FSD split, AraBERT achieved an accuracy of 93%, along with a macro F1 score of 0.84, while DistilBERT yielded lower accuracy (81%) and an F1 score of 0.67. Comparable trends were observed with the SSD and the USD; AraBERT maintained a stable F1 score of 0.84, indicating robustness, even when trained on pseudo-labelled data. DistilBERT's performance remained consistent but lower, with F1 scores ranging from 0.66 to 0.67 across all splits. These findings are summarised in Table 5.

Table 5: Sentiment classification results

| Dataset | Model | Accuracy | Macro Precision | Macro Recall | Macro F1 Score |
|---|---|---|---|---|---|
| FSD | AraBERT | 0.93 | 0.85 | 0.82 | 0.84 |
| FSD | DistilBERT | 0.81 | 0.66 | 0.67 | 0.67 |
| SSD | AraBERT | 0.92 | 0.83 | 0.86 | 0.84 |
| SSD | DistilBERT | 0.80 | 0.66 | 0.68 | 0.66 |
| USD | AraBERT | 0.92 | 0.84 | 0.84 | 0.84 |
| USD | DistilBERT | 0.81 | 0.67 | 0.69 | 0.67 |

### 4.4 Computational Efficiency

AraBERT consistently required longer training times than DistilBERT for both tasks. For topic classification, AraBERT averaged 18–19 minutes, while DistilBERT needed only 6–7 minutes. In sentiment classification, AraBERT's times ranged from 18 to 21 minutes, compared to DistilBERT's 6–8 minutes. The difference reflects AraBERT's higher model complexity.

## 5 DISCUSSION

The experimental results reveal several key insights into the use of LLM-based pseudo-labelling and the broader landscape of Arabic ABSA in the healthcare domain. Notably, the findings affirm the efficacy of pseudo-labelling using ChatGPT, which proved to be a reliable annotator for Arabic texts. The minimal performance gap – often less than 0.02 in the F1 score – between models trained on machine-generated labels and those trained on fully human-labelled data underscores the potential of this method in low-resource environments. This is especially promising for languages or domains in which manually annotated corpora are scarce, suggesting that with well-designed prompting strategies and robust LLMs, it is possible to generate training data of sufficient quality to support high-performing classifiers.

Despite the strength of pseudo-labels, human reviews still add measurable value, particularly in refining fine-grained taxonomies. The FSD consistently yielded slightly better performance, as human annotators were able to catch subtle misclassifications – especially those requiring nuanced domain knowledge or deeper contextual understanding. This indicates that a hybrid approach, in which LLM-generated labels are supplemented by selective human validation, offers a cost-effective compromise. In fact, our semi-supervised experiments demonstrated that reviewing only 50% of the data can yield results comparable to full manual annotation, highlighting the diminishing returns of exhaustive human labelling when high-quality pseudo-labels are available.

Model selection also played a significant role, particularly in relation to Arabic-specific language modelling. AraBERT consistently outperformed DistilBERT across tasks, emphasising the advantages of using models tailored to the Arabic language. Its deeper understanding of Arabic morphology, dialectal variations, and healthcare-specific terminology gave it a significant edge, especially in the more complex 17-class classification task. Nevertheless, the relatively strong performance of DistilBERT, despite being a general-purpose model, suggests its suitability for scenarios where slight trade-offs in accuracy are acceptable in exchange for faster training and inference times. This points to opportunities for future research to explore multilingual or domain-adapted models that strike a balance between performance and efficiency.

The impact of classification granularity was another important finding. Reducing the number of categories from 17 to 6 led to a notable improvement in classification performance. This supports the notion that the original taxonomy may have been too fine-grained or semantically overlapping, rendering consistent classification difficult. While detailed labels remain valuable for in-depth analysis, they also require more training data per class and are more prone to misclassification. A flexible taxonomy, as provided in the EHSAN dataset, allows researchers to select the level of detail appropriate for their specific objectives, and also encourages iterative taxonomy design – starting with detailed classes and clustering them based on confusion trends to achieve an optimal structure for machine learning applications.

Finally, a distinctive aspect of our dataset was the inclusion of model-generated rationales for each assigned label, aligning with the broader movement toward explainable AI in healthcare. Although we did not formally assess the impact of these explanations on user trust or debugging processes, qualitative feedback from the annotators suggested that they enhanced both confidence in the model's outputs and the interpretability of the annotation task. These rationales open up new possibilities, not only for enhancing annotation workflows but also for future model architectures that integrate explanation generation, such as multi-task learning setups or systems that warrant predictions alongside classification outputs.

Although the statistical analysis provided important insights into the reliability of the results, two considerations merit discussion. First, relying on only five runs may limit the precision of the confidence intervals, which could explain the noticeable variability observed in the confidence interval for the AraBERT model on the USD. In contrast, increasing the number of runs may lead to more stable estimates and narrower intervals. Second, the lack of statistical significance in some comparisons may be attributed to an underlying similarity in model behaviour across the different data settings, with only minor differences associated with the level of supervision. This is reflected in the comparison between the FSD and the USD in the AraBERT model, which did reach statistical significance.

In summary, the EHSAN framework proposed here offers a practical and scalable strategy for producing high-quality annotated datasets by effectively combining AI capabilities with human expertise. It enables fine-grained analysis of patient narratives in Arabic, demonstrating that, even in linguistically complex settings, modern NLP techniques can extract meaningful insights and support data-driven healthcare improvement. In terms of the study's limitations, considering that the present evaluation was limited to a specific healthcare domain and Saudi dialects, further work is needed to evaluate the framework's generalisability across other dialects and platforms. Another limitation is that our reliance on a single data source and language model may have introduced source- or model-specific biases.

## 6 CONCLUSION AND FUTURE WORK

This study introduced EHSAN, a comprehensive Arabic dataset and framework for the ABSA of healthcare reviews. By leveraging ChatGPT for initial labelling and integrating human validation in a tiered manner, we addressed key limitations in prior Arabic healthcare NLP efforts: data scarcity, lack of aspect specificity, and opaque model decisions. Our results show that an Arabic-focused model (AraBERT) can achieve robust performance in classifying topics and sentiments in patient reviews, even with minimal human supervision, thereby affirming the viability of LLM-based pseudo-labelling for building reliable datasets. In practical terms, the EHSAN framework can be integrated into real-time digital feedback monitoring systems within hospitals, enabling the tracking of patient sentiment on specific aspects, such as waiting times or billing. These insights can support timely administrative actions and guide data-driven quality improvement initiatives.

Several avenues emerged for future work from this study. First, cross-hospital and cross-region generalisation should be examined. Our models could be tested on patient reviews from different countries or healthcare settings to evaluate how well the insights generalise and to possibly expand EHSAN with more diverse data. In particular, since this study focused on healthcare reviews from Saudi Arabia, it would be valuable to examine the applicability of the findings across other Arabic-speaking countries, especially those with distinct dialects and healthcare delivery models. Future research could evaluate model performance and explanation quality when applied to reviews written in different regional varieties of Arabic, thereby assessing the robustness and adaptability of the proposed framework across the wider Arab world. Second, enhancing the prompting strategy for ChatGPT could further improve annotation consistency and reduce the observed error rate. Third, the integration of explanations into model training is an exciting direction: a model could be trained to not only predict labels but also to generate justification, which would move us closer to truly interpretable AI in this context. Fourth, fairness evaluation represents another important direction. Language models may exhibit biases in their outputs. In this study, we incorporated human reviewers to validate the model's outputs and reduce the likelihood of unintended bias, particularly in the classification of aspects and sentiments. Therefore, future studies could include a more systematic assessment of model-generated rationales across different dialects and aspect categories, examining whether these rationales contain implicit or unwarranted assumptions. In addition, analysing classification errors could help uncover potential bias patterns and address them through more precise and targeted prompting strategies to improve model fairness across varying contexts. Finally, enhancing dataset quality, such as increasing the number of annotated samples and ensuring balanced class distributions, may help reduce performance variability and lead to more statistically significant results across training configurations.

From a data-management standpoint, EHSAN already delivers a solid, production-ready pipeline; the next step is to amplify its reach. Three opportunities stand out for the IDEAS community to extend the platform's impact even further: 1) Streaming schema evolution: The aspect taxonomy must grow gracefully as hospitals introduce new services (e.g., a tele-ICU programme / remote intensive-care monitoring over video links). Supporting such additions on the fly, without interrupting dashboards or historical queries, will keep insights continuously up to date. 2) Explanation-provenance graphs: For every sentiment label the system assigns, it should be possible, instantly, to trace a path back to the exact ChatGPT prompt, the model's raw reply and any human corrections. Storing and querying this chain of evidence is non-trivial at scale. 3) Privacy-aware indexing: When we let users retrieve "reviews most similar to this one," the search index must guarantee k-anonymity so that rare disease mentions or unique patient details cannot be used to re-identify individuals.

In conclusion, the EHSAN dataset and approach fill a critical gap in Arabic healthcare text analysis by providing a high-resolution, explainable view of the patient experience. We hope that this work lays the groundwork for more patient-centric analytics in Arabic and other low-resource languages, and that the methods presented will inspire further innovations at the intersection of human expertise and AI-powered language understanding.

An anonymised version of the EHSAN dataset and the experimental code has been archived on Zenodo for perpetual access (https://doi.org/10.5281/zenodo.15418860).